\documentclass[conference]{IEEEtran}
\IEEEoverridecommandlockouts
\usepackage{cite}
\usepackage{amsmath,amssymb,amsfonts}
\usepackage{algorithmic}
\usepackage{comment}
\usepackage{graphicx}
\usepackage{placeins}
\usepackage{subcaption}
\usepackage{textcomp}
\usepackage{xcolor}
\usepackage{booktabs}
\usepackage{multirow}
\usepackage{siunitx}
\usepackage{hyperref}
\def\BibTeX{{\rm B\kern-.05em{\sc i\kern-.025em b}\kern-.08em
    T\kern-.1667em\lower.7ex\hbox{E}\kern-.125emX}}

\usepackage{ifthen}

\newcommand{\ccl}[1]{%
    \ifdim #1 pt < 0pt {\color{red} $#1$} \else $#1$ \fi
}

\newcommand{\cns}[1]{ {\color{gray} $#1$} }

\usepackage{pifont}
\newcommand{\cmark}{\ding{51}}%
\newcommand{\xmark}{\ding{55}}%

\newcommand{\VEC}[1]{\mathbf{#1}}
\newcommand{\opn}[1]{\operatorname{#1}}
\DeclareMathOperator{\E}{\mathbb{E}}
\DeclareMathOperator{\AMI}{AMI}
\DeclareMathOperator{\HH}{H}
\DeclareMathOperator{\II}{I}

\newcommand{\hide}[1]{}
\begin{document}

\title{Label-free Monitoring of Self-Supervised Learning Progress}

\author{\IEEEauthorblockN{Isaac Xu}
\IEEEauthorblockA{\textit{Faculty of Computer Science} \\
\textit{Dalhousie University}\\
Halifax, Canada \\
isaac.xu@dal.ca}
\href{https://orcid.org/0000-0003-4443-0582}{0000-0003-4443-0582}
\and
\IEEEauthorblockN{Scott Lowe}
\IEEEauthorblockA{\textit{Faculty of Computer Science} \\
\textit{Dalhousie University}\\
Halifax, Canada \\
scott.lowe@dal.ca\\
\href{https://orcid.org/0000-0002-5237-3867}{0000-0002-5237-3867}
}
\and
\IEEEauthorblockN{Thomas Trappenberg}
\IEEEauthorblockA{\textit{Faculty of Computer Science} \\
\textit{Dalhousie University}\\
Halifax, Canada \\
tt@cs.dal.ca \\
\href{https://orcid.org/0000-0002-6144-8963}{0000-0002-6144-8963}
}}

\maketitle

\begin{abstract}
Self-supervised learning (SSL) is an effective method for exploiting unlabelled data to learn a high-level embedding space that can be used for various downstream tasks.
However, existing methods to monitor the quality of the encoder --- either during training for one model or to compare several trained models --- still rely on access to annotated data.
When SSL methodologies are applied to new data domains, a sufficiently large labelled dataset may not always be available.
In this study, we propose several evaluation metrics which can be applied on the embeddings of unlabelled data and investigate their viability by comparing them to linear probe accuracy (a common metric which utilizes an annotated dataset).
In particular, we apply $k$-means clustering and measure the clustering quality with the silhouette score and clustering agreement.
We also measure the entropy of the embedding distribution.
We find that while the clusters did correspond better to the ground truth annotations as training of the network progressed, label-free clustering metrics correlated with the linear probe accuracy only when training with SSL methods SimCLR and MoCo-v2, but not with SimSiam.
Additionally, although entropy did not always have strong correlations with LP accuracy, this appears to be due to instability arising from early training, with the metric stabilizing and becoming more reliable at later stages of learning.
Furthermore, while entropy generally decreases as learning progresses, this trend reverses for SimSiam.
More research is required to establish the cause for this unexpected behaviour.
Lastly, we find that while clustering based approaches are likely only viable for same-architecture comparisons, entropy may be architecture-independent. 
\end{abstract}


\section{Introduction}
For many specialized fields seeking to deploy deep learning models, generating labels to produce viable datasets for supervised machine learning can be a costly process in terms of time and expertise.
Taking advantage of unlabelled data, recent self-supervised learning (SSL) methods have achieved state of the art performance as a means for extracting high-level features from complex inputs such as imagery \cite{simclr, mocov3, simsiam, byol}. 
These SSL methods have mainly been evaluated on labelled data.
In this work, we propose and evaluate metrics to monitor learning and the performance of the SSL models without annotations. 

In computer vision, an encoder model maps an input image from pixel-space to a lower dimensional representational space as an embedding vector which captures high-level contextual and semantic information in the image.
With supervised learning, the encoder is trained as part of the classification model. Then, through transfer learning, additional neural network ``head'' layers can be appended onto the encoder to interpret vectors in this embedding space for the purpose of other downstream tasks.

In SSL, the encoder can be trained with labels generated from the data itself via a ``pretext task'', meaning expensive human-annotation is not required.
An encoder trained with SSL can learn a more robust and generalizable embedding space than those generated from a supervised learning process \cite{ssl_robust, ssl_robustness_and_uncertainty}.
Recently, instance learning has been demonstrated to be a successful SSL pretext task.
In this method, the representational distance between independently augmented views of the same sample is minimized \cite{simsiam, byol}.
A subset of instance learning known as contrastive learning, also maximizes the distance between views of different samples \cite{simclr, mocov2}.

A common evaluation method for models prepared with SSL is to train a linear read-out layer on top of the encoder on a supervised classification task \cite{simsiam, byol, simclr, mocov2, mocov3}.
This process is referred to as a linear probe (LP).
Another increasingly popular method is to pass a dataset through the encoder and use a $k$-nearest neighbours (kNN) classifier \cite{mocov3, non-param_instance_discrimination}.
Since the distance between sample embeddings in the representational space carries semantic meaning, a sample's class can be predicted using the classes of its top $k$ nearest neighbours.
Although these methods only measure the performance on one downstream task (whole-frame classification), their performance is indicative of the utility of the embedding space on other tasks.
However, these methods still require labelled test datasets, which can be challenging to acquire.

\begin{figure*}[t]
\begin{subfigure}[b]{0.315\linewidth}
\centering
\includegraphics[width=\textwidth]{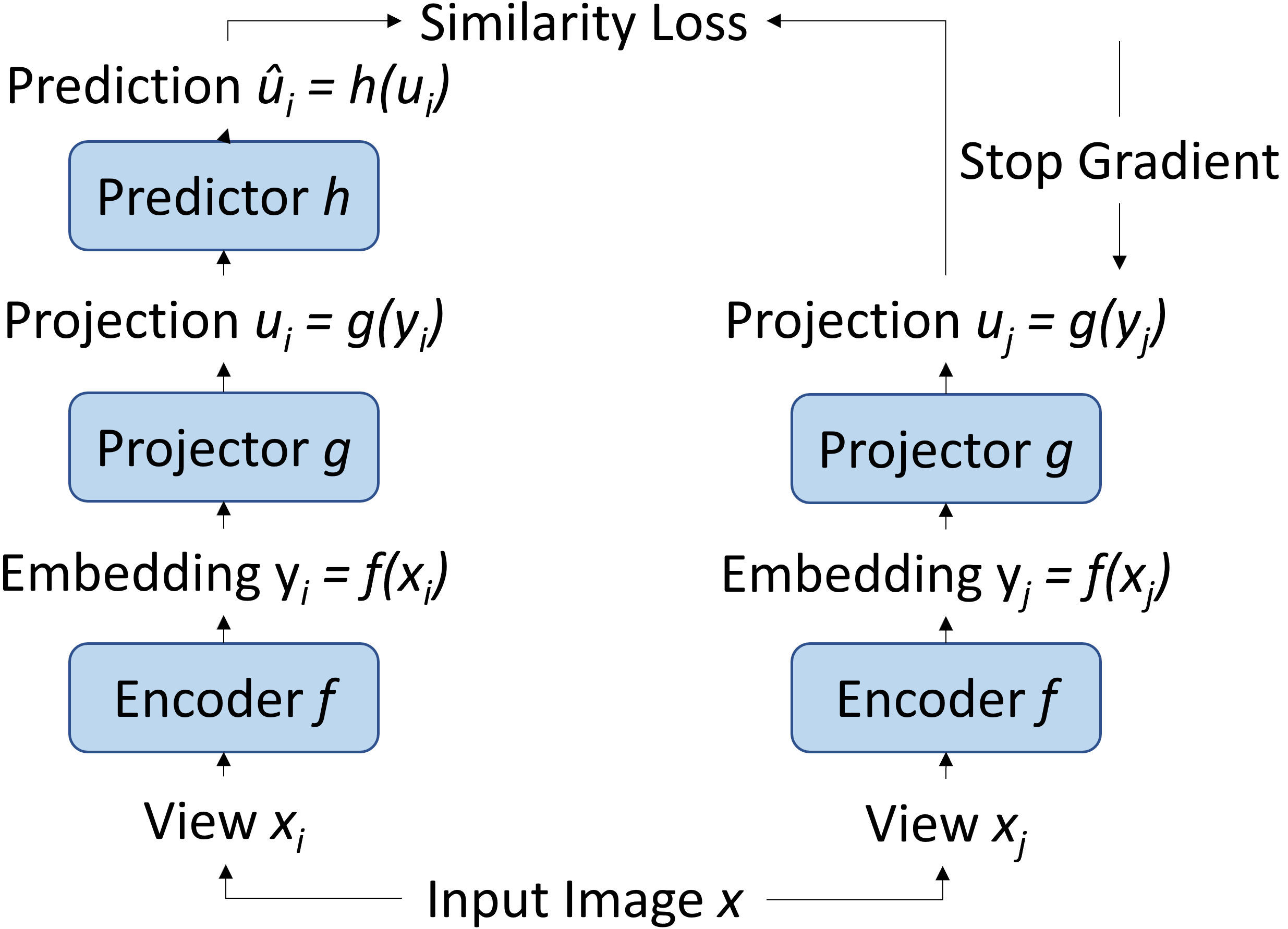}
\caption{SimSiam.}
\label{fig:simsiam-arch}
\end{subfigure}
\hspace{0.345cm}
\begin{subfigure}[b]{0.315\linewidth}
\centering
\includegraphics[width=\textwidth]{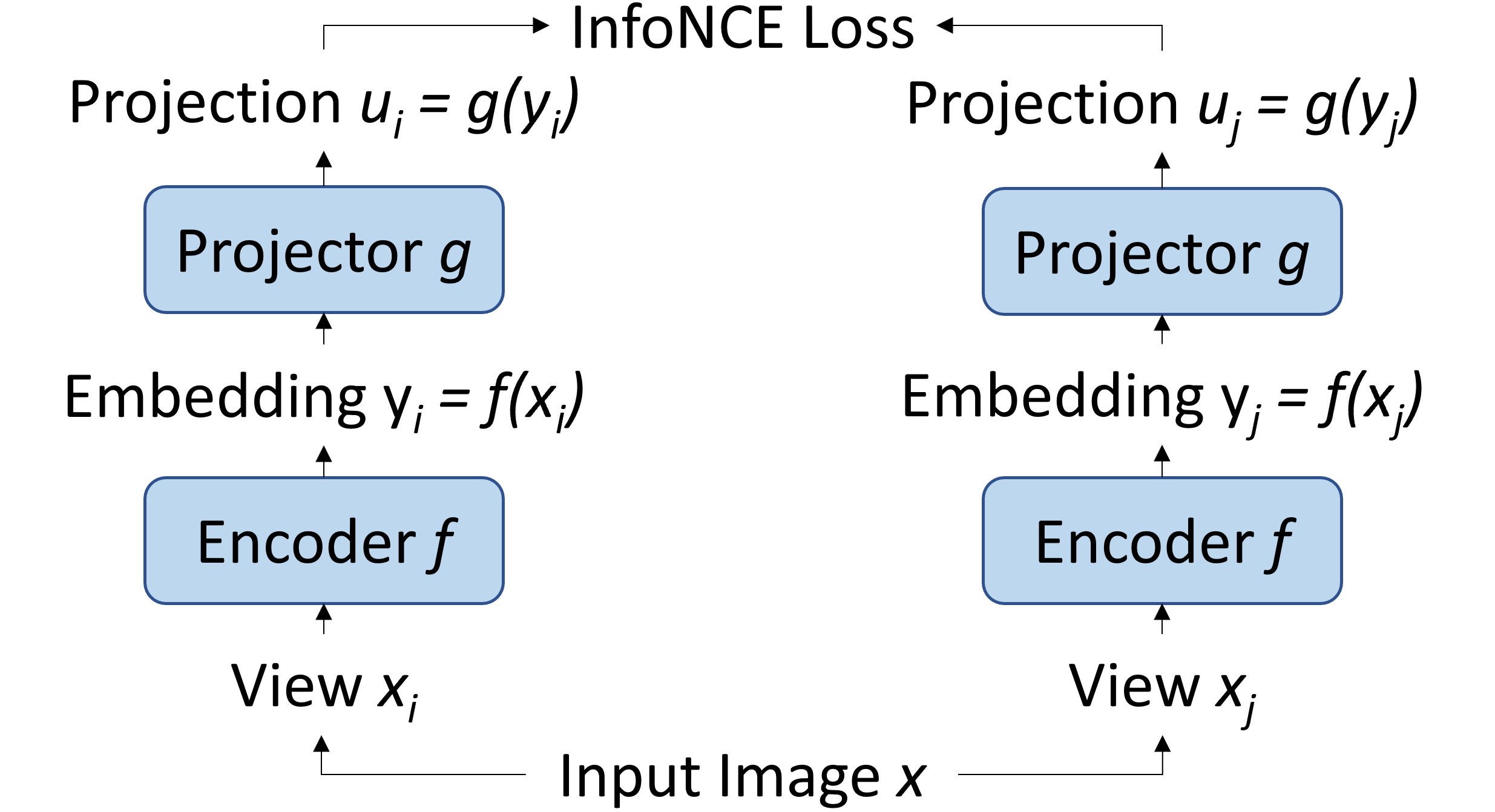}
\caption{SimCLR.}
\label{fig:simclr-arch}
\end{subfigure}
\hspace{0.345cm}
\begin{subfigure}[b]{0.315\linewidth}
\centering
\includegraphics[width=\textwidth]{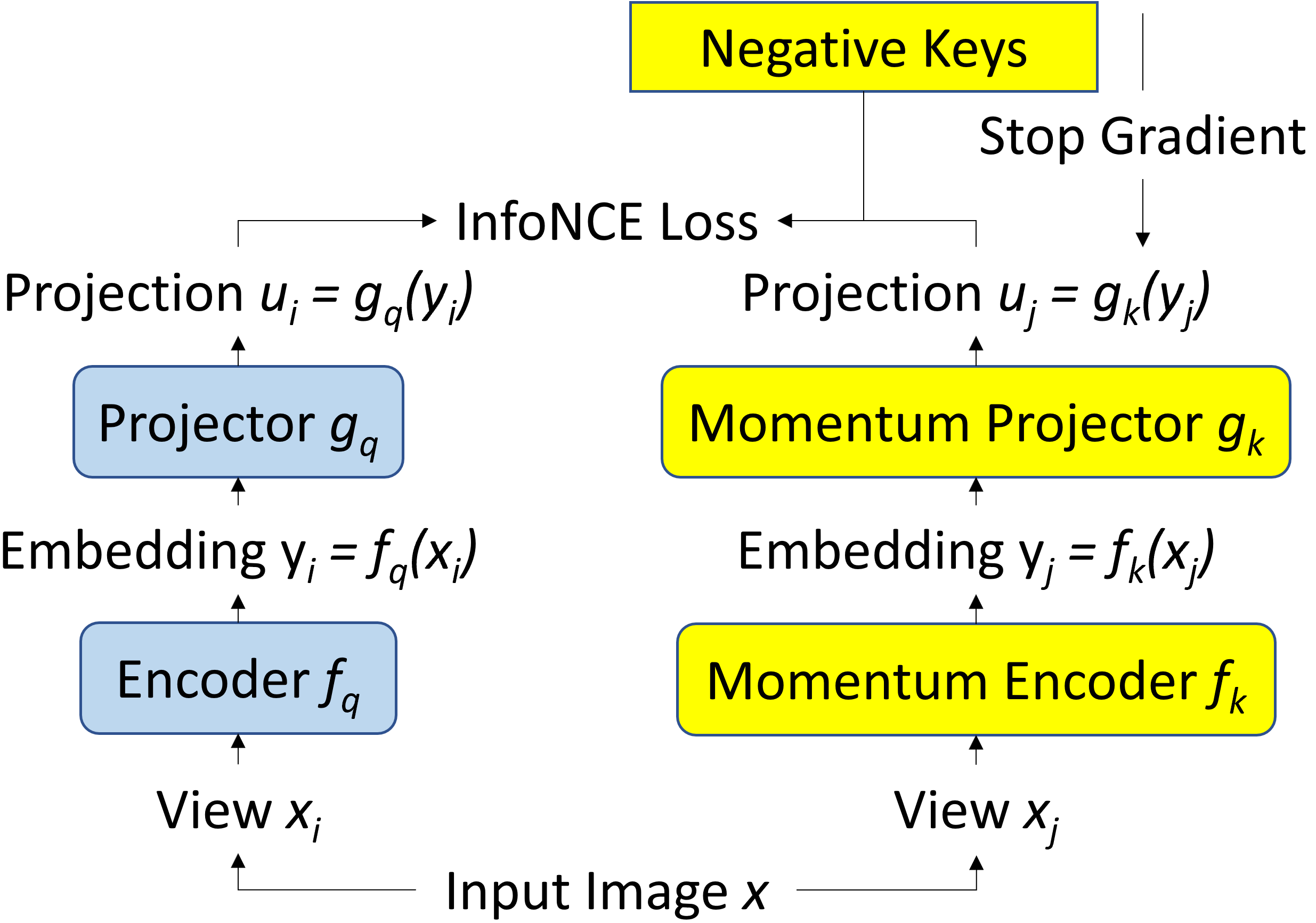}
\caption{MoCo-v2.}
\label{fig:mocov2-arch}
\end{subfigure}
\caption{
Overview of SSL methodologies.
For a given image, $x$, two views of the image ($x_i$ and $x_j$) are created with randomly sampled augmentations.
In SimSiam (\subref{fig:simsiam-arch}) and SimCLR (\subref{fig:simclr-arch}), the two views are passed through the same encoder, $f$, and projector, $g$, layers; in MoCo-v2 (\subref{fig:mocov2-arch}), view $x_j$ passes through a moving average model instead.
For SimSiam, the loss is the similarity between the output of a predictor head on the $x_i$ branch and the projection of $x_j$; whereas for SimCLR and MoCo-v2, InfoNCE is used with negative samples drawn from the batch (SimCLR) or the memory queue (MoCo).
In SimSiam and MoCo-v2, a stop-gradient is applied to prevent the loss returning down the $x_j$ branch of the network.
}
\label{fig:ssl-arch}
\end{figure*}

Our proposal for monitoring learning progress without labels is based on the conjecture that due to the semantic meaning in learned embedding space, similar samples should increasingly group together over learning.
Hence, a core part of our evaluation approach relies upon employing $k$-means clustering on sample embeddings and characterization of the clusters using traditional clustering metrics.
We also look to the agreement between two independent $k$-means clustering attempts to provide an intuition for clustering consistency.
We expect that the more well-formed ground truth clusters are, the more consistently $k$-means would capture these clusters.
Lastly, we measure the entropy of the data embeddings, as we hypothesise that entropy will decrease during training when we progress from an initial high-dispersion and low-modal distribution to a more densely packed higher-modal distribution. 

We compare these results to LP accuracy, the current standard measure of model quality.
A stronger correlation with LP accuracy indicates greater viability for the label-free metric.
We apply our proposed metrics on three SSL techniques: SimSiam \cite{simsiam}, SimCLR \cite{simclr}, and MoCo-v2~\cite{mocov2}.
Our experiments were performed on CIFAR-10 and CIFAR-100~\cite{cifar10} datasets.

Additionally, we applied these metrics to a variety of architectures pre-trained with supervised learning on ImageNet~\cite{imagenet}.
These were ResNet \cite{resnet}, EfficientNet \cite{efficientnet}, and DenseNet \cite{densenet} models provided in torchvision~\cite{pytorch}.
Here, we examine if these embedding measures are capable of distinguishing between the quality of models using different architectures.

\section{Background and Related Work}

\subsection{Instance learning}
The SSL methods we use employ strong augmentations to create a pretext task.
Each sample from a particular dataset is augmented twice independently, producing two ``views'' of the sample. 
The encoder is motivated to generate embeddings which are robust against the augmentation operation, such that the independent views for the same sample have similar embeddings.

\subsubsection{SimSiam}
With the SimSiam training configuration \cite{simsiam}, both views ($i$ and $j$) pass through the encoder, plus a multi-layer perceptron (MLP) section dubbed the projector.
One view passes through another MLP dubbed the predictor.
The network is trained to minimise the distance between the projected representation of one view, $\VEC{u}_i$, and the prediction generated from the second view, $\hat{\VEC{u}}_j$, as given by
\begin{equation}
    L_{i,j} = -\sqrt{\operatorname{d_{cos}}(\VEC{u}_i,\hat{\VEC{u}}_j) + \operatorname{d_{cos}}(\VEC{u}_j,\hat{\VEC{u}}_i)}
,\end{equation}
where $\operatorname{d_{cos}}(\cdot,\cdot)$ is the cosine similarity distance measure.
In order to prevent collapse to a trivial solution, updates to minimize this loss propagate back only through the prediction branch.
We can interpret this process as tasking the model to predict the average embedding (in the projector space) over all views of a sample, when presented with any one view of that sample.
The SimSiam method is displayed in \autoref{fig:simsiam-arch}.

\subsubsection{SimCLR}
Meanwhile, the SimCLR approach uses contrastive learning, as seen in \autoref{fig:simclr-arch}.
Views of other same batch samples are used as negative views and the network must learn embeddings such that two views of the same sample (positive views) are close together, whilst negative views are far apart.
In the SimCLR formulation, the predictor head is not used and we only need consider the projected view, $\VEC{u}$.
For a given pair of positive views, $i$ and $j$, in a minibatch of $N$ samples, the loss is
\begin{equation}
    L_{i, j} = -\log{\frac{\exp{(\operatorname{d_{cos}}(\VEC{u}_i, \VEC{u}_j)/{\tau})}}{\sum_{k=1}^{2N}(1-\opn{\delta}_{ik})\exp{(\operatorname{d_{cos}}(\VEC{u}_i, \VEC{u}_k)/{\tau})}}}
,\end{equation}
where $\opn{\delta}$ is the Kronecker delta function and $\tau$ is a temperature scaling factor~\cite{simclr}. 
This loss is also referred to as the normalized temperature-scaled cross entropy loss \cite{simclr} or as InfoNCE~\cite{infonce}.

\subsubsection{MoCo-v2}
The final method MoCo-v2, seen in \autoref{fig:mocov2-arch}, shares the InfoNCE loss function with SimCLR.
The main difference between the two methods lies in the use of a momentum or ``key'' encoder and a memory queue for the projections of negative views, referred to as negative keys.
Here, the main encoder is referred to as a query encoder. 
The key encoder parameters are updated as the exponential moving average (EMA) of the query encoder's parameters at every batch.
The views processed by the key encoder are then added to the queue of negative keys.
The objective is then to compare the ``query'' view projections against the positive keys (same sample view projections) and the negative keys.

\subsection{Extrinsic metrics}
Extrinsic metrics are measures of clustering quality based on an external reference.
In this work, we use the mutual information \cite{shannon_mutual_info} between the cluster labels generated by $k$-means and the ground truth class labels, acting as a benchmark to compare label-free metrics against.
We can imagine the class and cluster labels to be two discrete random variables.
For discrete random variables $X$ and $Y$, their mutual information is defined as
\begin{equation}
    \II(X;Y) = \sum_{y \in Y} \sum_{x \in X} P(x,y) \log \left(\frac{P(x,y)}{P(x) P(y)}\right)
,\end{equation}
where $P(x,y)$ denotes the joint probability distribution for $X$ and $Y$, while $P(x)$ and $P(y)$ denotes their respective marginal distributions.
Intuitively, we can conceptualize mutual information as a measure of how much information we can obtain about $Y$ upon observing $X$ and vice versa.

For implementation, we use the adjusted mutual information ($\AMI$) score from the scikit-learn library \cite{scikit-learn}. 
The AMI corrects for the chance level of mutual information which would be measured given a certain finite number of samples. 
The AMI between two discrete random variables $X$ and $Y$ is defined as
\begin{equation}
\AMI(X;Y) = \frac{\II(X;Y)-\E[\II(X^*;Y^*)]}{(\HH(X)+\HH(Y))/2-\E[\II(X^*;Y^*)]}
,\end{equation}
where $\HH$ is entropy and the expected information is taken over a hypergeometric model of $X$ and $Y$. 

\subsection{Intrinsic measures}
Intrinsic measures of clustering quality consider properties of the clusters, such as inter-cluster and intra-cluster distances.
The silhouette score is one such measure \cite{silhouette_score}. 
For each sample, $i$, the silhouette score is defined as
\begin{equation}
    S_i = \frac{b_i-a_i}{\max(a_i,b_i)}
,\end{equation}
where $b_i$ is the average inter-cluster distance between sample $i$ and the nearest neighbouring cluster's samples, while $a_i$ is the average intra-cluster distance from sample $i$ to other same cluster samples.
The individual silhouette scores are averaged over all samples to provide an overall silhouette score.

Our work differs from deep clustering \cite{deepcluster_survey} in that rather than incorporating clustering into the training process, we are looking to evaluate a trained model (of varying degrees) using clustering.
Furthermore, a major goal is to dissociate model evaluation from label schemes and therefore we cannot rely upon the use of extrinsic measures which take into account ground truths.
We use such extrinsic measures only as a means of comparison against investigated label-free methods.

\section{Methods}



\subsection{Datasets}
Our networks were trained on either the CIFAR-10 or CIFAR-100 dataset \cite{cifar10}.
These datasets consist of $32\times32$ pixel natural RGB images, with either 10 or 100 classes, respectively.
All training was performed on the training partition and representations were evaluated using the test partition of these datasets.

\subsection{Self-supervised learning}
We used the modified ResNet-18 \cite{resnet} backbone from SimCLR \cite{simclr} for CIFAR-10 images, with a 512 dimensional representation space.
During SSL, we augmented the images to create pairs of views using the CIFAR-10 augmentation stack from SimCLR \cite{simclr}.
When training SimSiam, we added a three layer projector and a two layer predictor, with a width of 2048 throughout except for the bottleneck layer of the predictor, which had a width of 512.
For SimCLR, we instead added a two-layer projector, with hidden dimension 2048 and output dimension 128.
These are also the same dimensions we use for the MoCo architectures.

The networks were trained with SSL using stochastic gradient descent (SGD) using a one-cycle learning rate schedule \cite{one-cycle} with cyclic momentum from 0.85 to 0.95.
For SimSiam, the peak learning rate $\eta=0.06$, weight decay $\lambda=\num{5e-4}$, and we trained the network for 800 epochs.
For SimCLR, $\eta=0.5$, $\lambda=\num{1e-4}$, temperature $\tau=0.5$, and we trained the network for 1000~epochs.
Lastly, for MoCo-v2, $\eta=0.06$, $\lambda=\num{5e-4}$, $\tau=0.1$, queue length was 4096, EMA multiplier $m=0.99$, and the network was trained for 800~epochs.

\subsection{Representation evaluation}
Every 20 epochs of the SSL process, we created a checkpoint of the model, referred to as a ``milestone''.
For each milestone, we passed the test partition samples through the encoder backbone to generate a set of 512-d embedding vectors, $Z_{512}$.
Typical clustering methods do not work well on large representation spaces \cite{curse_of_dims}, so we reduced $Z_{512}$ down to a 3-d space, using uniform manifold approximation and projection (UMAP) \cite{umap} with $n=50$ neighbours.
We then clustered the 3-d UMAP projections of the embedding vectors $Z_3$, using $k$-means with the cosine distance metric and $k_1=10$ or $k_1=100$ as per the number of annotated classes in the dataset.
The cluster labels generated from this clustering are referred to as $C_1$. 

We evaluated the quality of $C_1$ by measuring the amount of information about the ground truth labels $C_{\text{GT}}$, contained in $C_1$ using $\AMI(C_1;C_{\text{GT}})$.
This serves as a baseline to compare our other metrics against.

We defined $S_1$ as the silhouette score for the clusters $C_1$, evaluated by Euclidean distance for the original embedding vectors $Z_{512}$.
We similarly defined $S_{\text{GT}}$ as the silhouette score of $C_{\text{GT}}$, also for $Z_{512}$.
This measurement provides an upper-bound on the \emph{utility} of $S_1$ --- that which could be obtained with ``perfect'' cluster assignments.

Finally, we used $k$-means again to generate a second set of clusters, $C_2$, with double the classes: $k_2=2k_1$.
We then defined the \textit{clustering agreement} as the adjusted mutual information between these two sets of clusters, $\AMI(C_1;C_2)$.

\begin{table*}[htb]
\caption{Pearson correlation between performance metrics and linear probe accuracy. We display correlation scores both with (w/ init) and without (w/o) the network initialization (i.e. before training begins) milestone included in the trend. Grey: no significant correlation ($p<0.05$). Black: positively correlated. Red: negatively correlated.}
\label{tab:ssl}
\centerline{
  \scalebox{0.96}{
\begin{tabular}{lcrrrrrrrrrrrr}
\toprule
& &\multicolumn{4}{c}{{SimSiam}} &\multicolumn{4}{c}{{SimCLR}} &\multicolumn{4}{c}{{MoCo-v2}} \\
\cmidrule(lr){3-6} \cmidrule(lr){7-10} \cmidrule(lr){11-14}
& &\multicolumn{2}{c}{{CIFAR-10}} &\multicolumn{2}{c}{{CIFAR-100}}&\multicolumn{2}{c}{{CIFAR-10}} &\multicolumn{2}{c}{{CIFAR-100}}&\multicolumn{2}{c}{{CIFAR-10}} &\multicolumn{2}{c}{{CIFAR-100}}\\
\cmidrule(lr){3-4} \cmidrule(lr){5-6} \cmidrule(lr){7-8} \cmidrule(lr){9-10}
\cmidrule(lr){11-12} \cmidrule(lr){13-14} 
{Metric}
    & Label-free
    & \multicolumn{1}{c}{w/ init}
    & \multicolumn{1}{c}{w/o}
    & \multicolumn{1}{c}{w/ init}
    & \multicolumn{1}{c}{w/o}
    & \multicolumn{1}{c}{w/ init}
    & \multicolumn{1}{c}{w/o}
    & \multicolumn{1}{c}{w/ init}
    & \multicolumn{1}{c}{w/o}
    & \multicolumn{1}{c}{w/ init}
    & \multicolumn{1}{c}{w/o}
    & \multicolumn{1}{c}{w/ init}
    & \multicolumn{1}{c}{w/o}
\\
\midrule
$\AMI({C_1, C_\text{GT}})$   & \xmark &  \ccl{0.97} &  \ccl{0.96} &  \ccl{0.98} &  \ccl{0.98} &  \ccl{0.93} &  \ccl{0.97} &  \ccl{0.93} &  \ccl{0.97} &  \ccl{0.92} &  \ccl{0.95} &  \ccl{0.88} &  \ccl{0.91} \\
$\AMI({C_1,C_2})$            & \cmark &  \cns{0.00} & \cns{-0.21} & \ccl{-0.62} & \ccl{-0.71} &  \ccl{0.71} &  \ccl{0.91} &  \ccl{0.76} &  \ccl{0.90} &  \ccl{0.64} &  \ccl{0.80} &  \ccl{0.47} &  \ccl{0.61} \\
$S_{\text{GT}}$              & \xmark &  \ccl{0.93} &  \ccl{0.97} &  \ccl{0.94} &  \ccl{0.96} &  \ccl{0.89} &  \ccl{0.62} &  \ccl{0.95} &  \ccl{0.91} &  \ccl{0.37} & \cns{-0.29} &  \ccl{0.51} & \cns{-0.11} \\
$S_1$                        & \cmark & \cns{0.05} & \ccl{-0.59} &  \cns{0.16} & \ccl{-0.57} &  \ccl{0.59} & \cns{-0.07} &  \ccl{0.28} & \ccl{-0.78} & \cns{-0.09} & \ccl{-0.78} & \ccl{-0.47} & \ccl{-0.85} \\
$\HH(Z_3)$                   & \cmark &  \ccl{0.82} &  \ccl{0.92} &  \ccl{0.86} &  \ccl{0.87} & \ccl{-0.57} & \ccl{-0.83} &  \cns{0.18} & \cns{-0.20} & \cns{-0.26} & \ccl{-0.62} &  \ccl{0.60} &  \ccl{0.47} \\
\bottomrule
\end{tabular}
}
}
\end{table*}

\begin{table*}[htb]
\caption{Pearson correlation between performance metrics and linear probe accuracy when transferring models pre-trained on ImageNet to CIFAR-10 or CIFAR-100. Correlations were measured across pre-trained models of the same architecture (ResNet, EfficientNet, or DenseNet), but different sizes. Grey: no significant correlation ($p<0.05$). Black: $+$ve correlation. Red: $-$ve.}
\label{tab:sl}
\centerline{
\begin{tabular}{lcrrrrrrrr}
\toprule
& &\multicolumn{2}{c}{{ResNet}}
&\multicolumn{2}{c}{{DenseNet}}
&\multicolumn{2}{c}{{EfficientNet}}
&\multicolumn{2}{c}{{Overall}}\\
\cmidrule(lr){3-4} \cmidrule(lr){5-6} \cmidrule(lr){7-8} \cmidrule(lr){9-10}
{Metric}
    & Label-free
    & CIFAR-10
    & CIFAR-100
    & CIFAR-10
    & CIFAR-100
    & CIFAR-10
    & CIFAR-100
    & CIFAR-10
    & CIFAR-100\\
\midrule
$\AMI({C_1, C_\text{GT}})$  & \xmark &  \cns{0.86} &  \ccl{0.92} &  \cns{0.73} &  \cns{0.69} &  \ccl{0.80} &  \ccl{0.72} &  \cns{0.20} &  \cns{0.38} \\
$\AMI({C_1,C_2})$           & \cmark &  \cns{0.83} &  \ccl{0.96} & \cns{-0.67} &  \cns{0.32} &  \cns{0.38} &  \cns{0.45} &  \cns{0.03} &  \cns{0.23} \\
$S_{\text{GT}}$             & \xmark &  \ccl{0.95} &  \ccl{0.97} & \cns{-0.15} &  \cns{0.10} &  \cns{0.24} &  \ccl{0.83} &  \cns{0.16} &  \cns{0.21} \\
$S_1$                       & \cmark &  \ccl{0.95} &  \ccl{0.99} &  \cns{0.44} & \cns{-0.06} & \cns{-0.16} & \cns{-0.12} & \cns{-0.06} &  \cns{0.13} \\
$\HH(Z_3)$                  & \cmark & \cns{-0.19} & \cns{-0.87} &  \cns{0.75} & \cns{-0.89} & \ccl{-0.85} & \cns{-0.54} & \cns{-0.34} & \ccl{-0.52} \\
\bottomrule
\end{tabular}
}
\end{table*}

To measure the entropy of the embedding space, we take the $Z_3$ vectors and bin the values along each dimension to yield a 3-d histogram.
The bin width is set separately for each dimension as $l_i=0.4\,\sigma_i$, where $\sigma_i$ is the standard deviation of the $Z_3$ vectors in dimension $i$.
The 3-d histogram bin counts are divided by the total number of test samples, to yield an empirically observed probability distribution, from which we measure the entropy.

\subsection{Pre-trained models}
We loaded pre-trained models provided in torchvision~\cite{pytorch}, which had been trained on ImageNet-1k.
We used models with ResNet, DenseNet, and EfficientNet architectures of varying sizes: ResNet 18, 34, 50, 101, and 152; DenseNet 121, 161, 169, and 201; EfficientNet (v1) sizes b0 through b7.
The same methodology as described above was applied, using the pre-trained model as a (frozen) encoder, with the following changes.
(1) CIFAR-10 and -100 images were upscaled to the same resolution as that which the model was trained on.
(2) Due to the significantly larger image resolutions for the bigger EfficientNet models, we reduced all batch sizes to 48. 
(3) The maximum learning rate was reduced to 0.003 for LPs.
(4) The entropy bin width was increased to $l_i=0.8\,\sigma_i$, because the distribution in the representation space was found to be up to twice as large for SL pre-trained models when compared with SSL.

\begin{figure}[htb]
\begin{minipage}[b]{0.95\linewidth}
\centering
\includegraphics[width=0.95\textwidth]{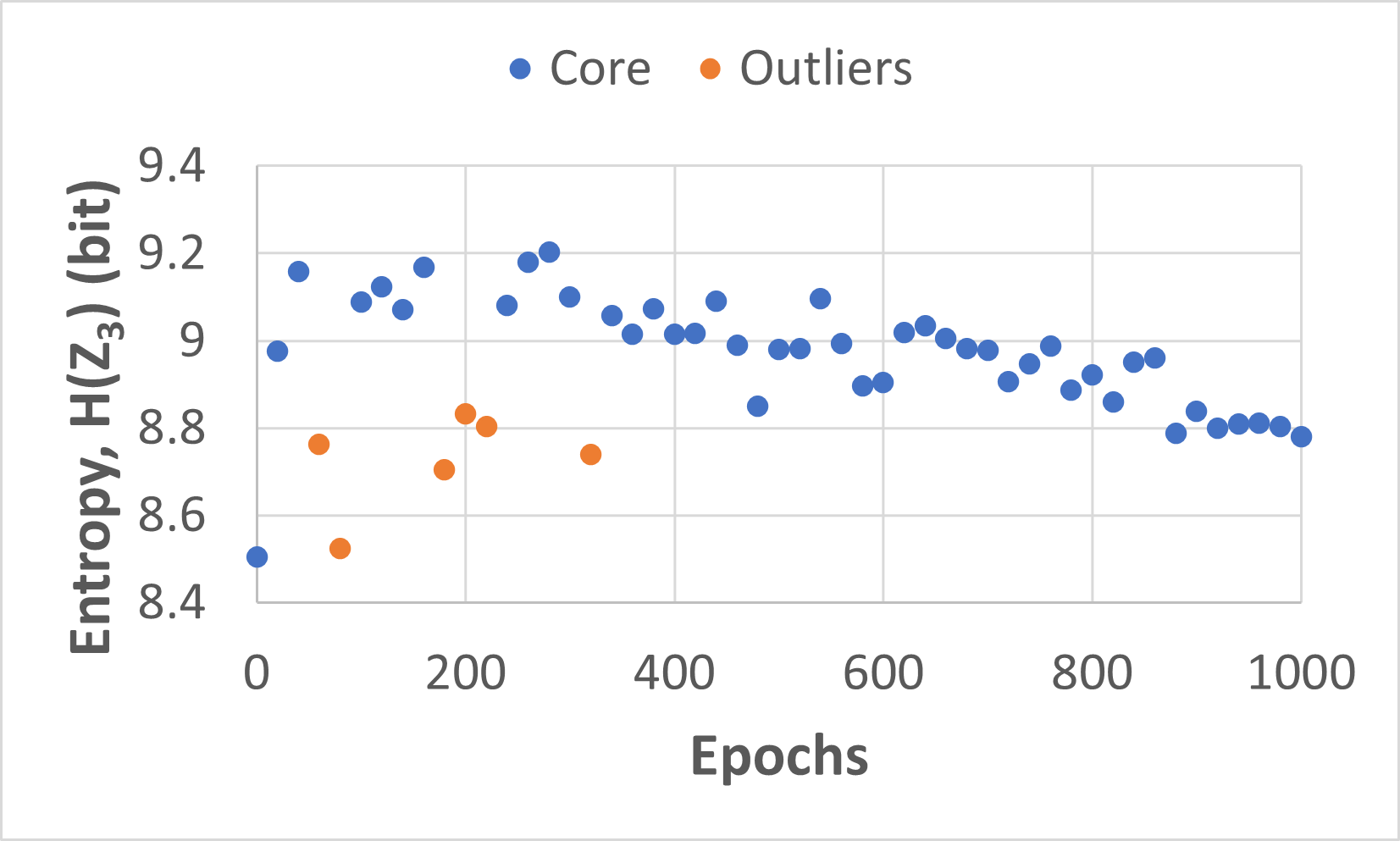}
\caption{Entropy progression w.r.t. training for SimCLR on CIFAR-100. We highlight milestones with outlying samples in the embedding space (orange).
}
\label{fig:simclr-entropy-cifar100}
\end{minipage}
\end{figure}

\begin{figure*}[htb]
\begin{minipage}[b]{0.475\linewidth}
\centering
\includegraphics[width=\textwidth]{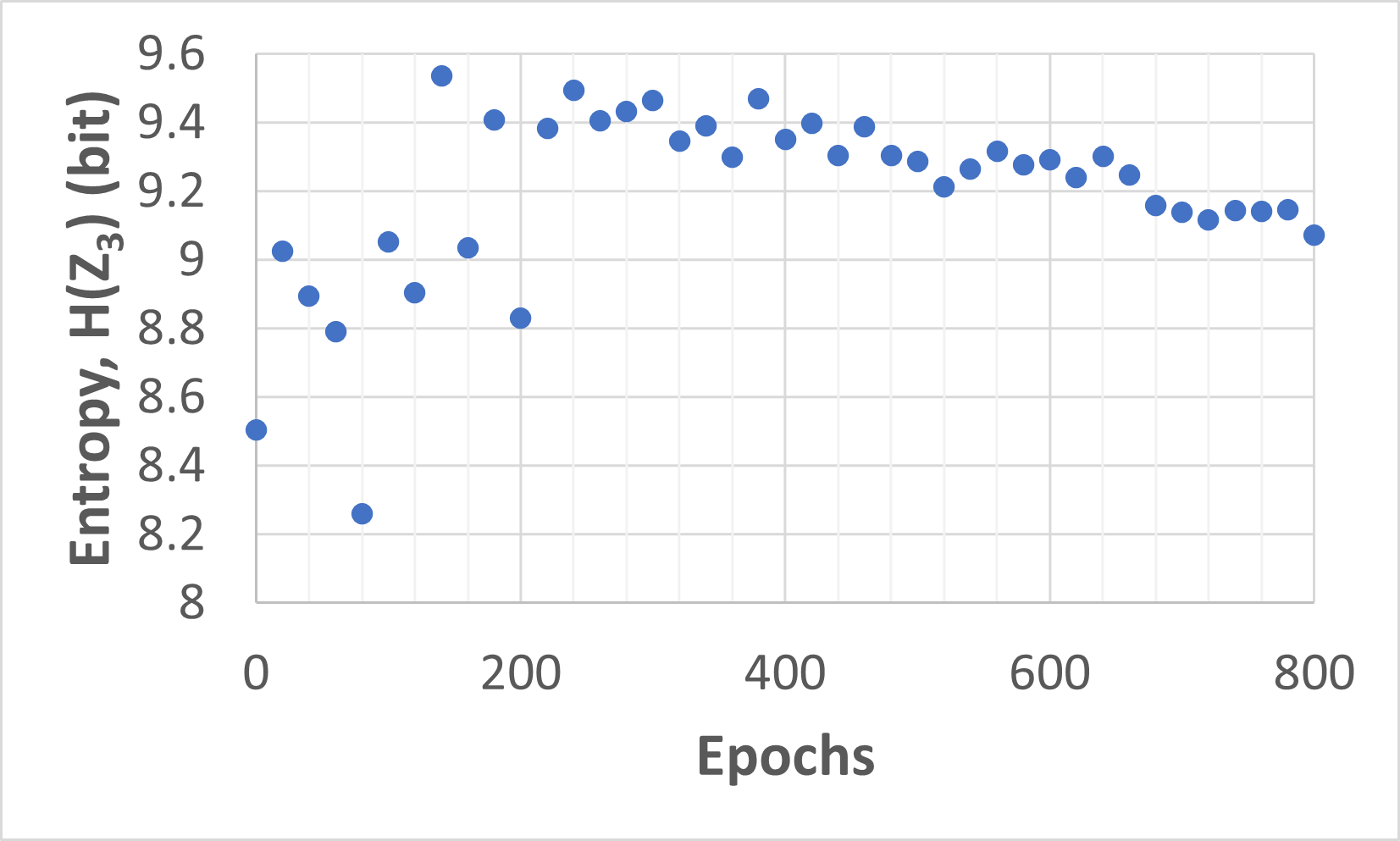}
\caption{\raggedright Entropy progression w.r.t training for MoCo-v2 on CIFAR-100.}
\label{fig:mocov2-entropy-cifar100}
\end{minipage}
\hspace{0.5cm}
\begin{minipage}[b]{0.475\linewidth}
\centering
\includegraphics[width=\textwidth]{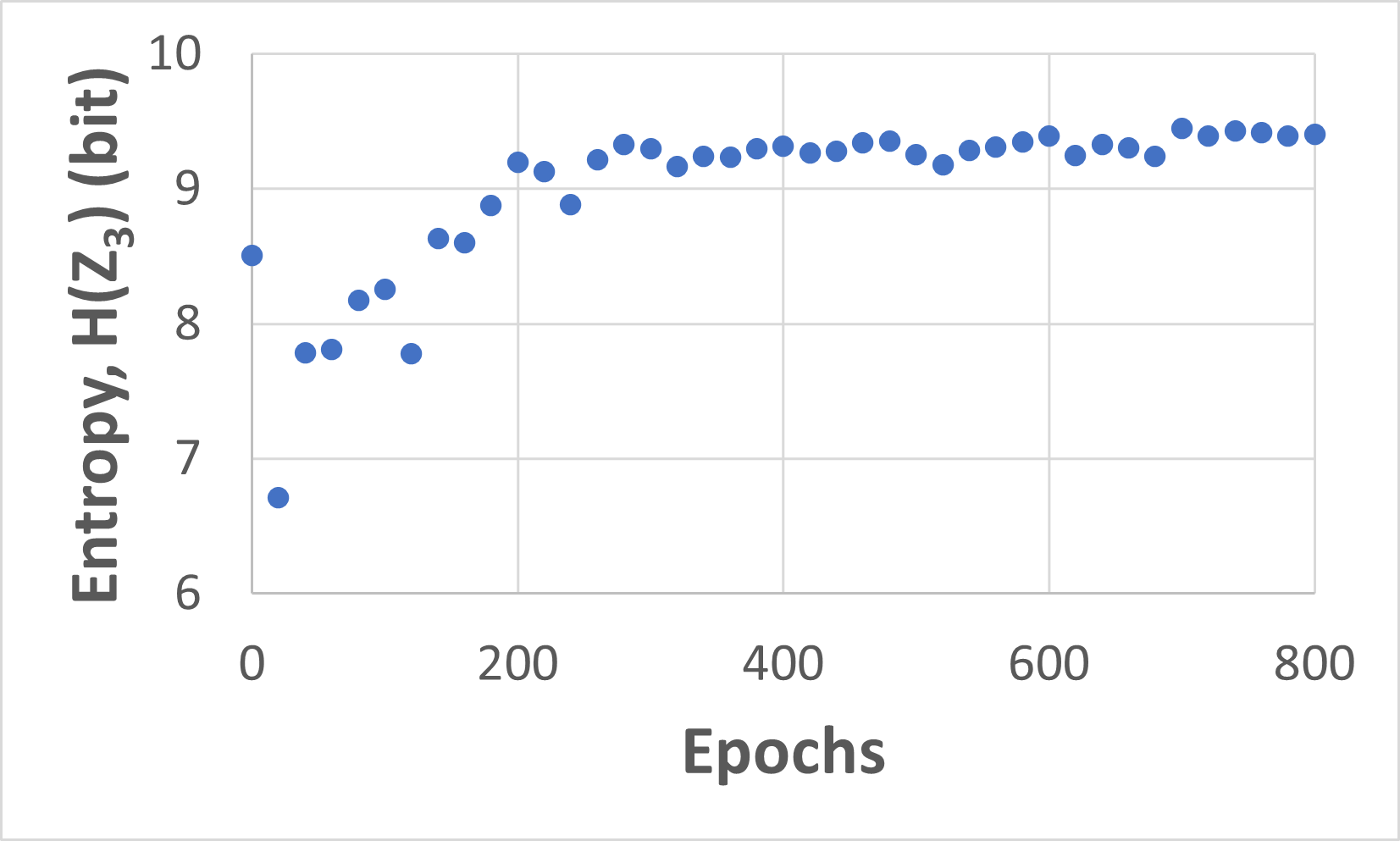}
\caption{\raggedright Entropy progression w.r.t training for SimSiam on CIFAR-100.}
\label{fig:simsiam-entropy-cifar100}
\end{minipage}
\end{figure*}

\subsection{Linear probe}
For each milestone, we extracted the encoder from the network and froze its weights.
We added a linear layer on top of the encoder and used the training partition to learn a linear mapping from the embedding space to the target labels.
The linear layer was trained with the Adam optimizer \cite{adam}, for 20 epochs, using a one-cycle learning rate schedule with peak learning rate of 0.08, cyclic momentum from 0.85 to 0.95, and weight decay of \num{1e-4}.
During LP training, we used the augmentations from the CIFAR-10 policy discovered by AutoAugment~\cite{autoaug}.

\section{Results}
We measured the Pearson correlation over training milestones between our metrics and the LP accuracy, the results of which are shown in \autoref{tab:ssl}.
Similarly, we measure the correlation for the torchvision pre-trained models separately for each architecture type and in a general ``overall'' case, presented in \autoref{tab:sl}.
The overall case is intended to be a measure of how feasible a metric may be for cross-architecture comparison.
The notations used are as in the previous section, with subscripts $\text{GT}$ indicating ground truth labels and $1$ indicating clusters $C_1$.
Only metrics pertaining to $C_1$ require $k$-means clustering to be performed.

We observe that $\AMI({C_1, C_\text{GT}})$ correlates strongly with LP accuracy throughout all combinations of methods and datasets. 
The metric also performs well across the pre-trained models but could not be extended to cross-architecture evaluation, as evidenced by its weak correlation in the overall case.  
This result demonstrates that clustering is indeed able to progressively pick out ground truth classes as clusters, reflective of learning progression with the likely caveat of being limited to same-architecture comparisons.
Although the silhouette score $S_{\text{GT}}$ remains near zero ($|S_{\text{GT}}|<0.05$) and assigns a poor score to the ground truth interpreted as clusters, we find it is well correlated with the LP measurements in SimSiam and SimCLR cases. 
However, it appears to be inconsistent for MoCo-v2 as well as potentially for EfficientNet and DenseNet architectures, suggesting that the embedding clustering shapes even when representing ground truth, may not progress in a manner reflective of learning progression.

Our results show that the label-free metrics we investigated only present weak correlations with the label-based scores on average.
More specifically, the silhouette score $S_1$ did not correlate well with LP accuracy. As is the case with its ground truth counterpart, it also assigned poor scores to minimally separated clusters.
We also found that there could be a large change in the correlation if the initial network state is dropped from the correlates.
Due to the non-linearity of learning progression, equally spacing out milestones in terms of epochs can lead to a relatively lower number of data points reflective of early training. 
As such, early training progression which may differ in nature from later training progression, may not be as sufficiently captured.

Although clustering agreement $\AMI({C_1,C_2})$, was correlated with LP accuracy for SimCLR and to a lesser extent, MoCo-v2, it was either not correlated or inversely correlated when training with SimSiam. 
This observation may be tied to the unexpected result that entropy increases for SimSiam but decreases for SimCLR.
We hypothesis that this may be because SimSiam is more susceptible to dimensional collapse \cite{feature_decorrelation, understanding_dim_collapse_in_ssl}.
In future work, this could be confirmed by observing the singular value spectra of the projector and predictor embedding spaces.

We expected the entropy of the embeddings to be inversely correlated with LP accuracy, since a distribution of embedding vectors which is more tightly clustered will be associated with lower entropy.
For SSL, this was only the case for SimCLR and MoCo-v2 on CIFAR-10.
Investigating the distribution of CIFAR-100 $Z_3$ embeddings in early milestones, we discovered that SimCLR had outlier embeddings (at a large distance away from the rest of the samples) which caused a lower entropy measurement, due to its sensitivity to the size of the space spanned by the embeddings.
As shown in \autoref{fig:simclr-entropy-cifar100}, these outliers occur sporadically during the early stages of training and the trend for entropy during training is otherwise consistent.
In \autoref{fig:mocov2-entropy-cifar100}, we observe a similar issue for MoCo-v2 training on CIFAR-100, where due to their low entropy measure, the early milestones cause the metric to be positively correlated to LP accuracy. 
It is important to note however, that this is not the case for SimSiam's positive entropy correlation, where the relationship is largely consistently positive throughout training, as can be seen in \autoref{fig:simsiam-entropy-cifar100}.

\begin{figure*}[htb]
\begin{minipage}[b]{0.475\linewidth}
\centering
\includegraphics[width=\textwidth]{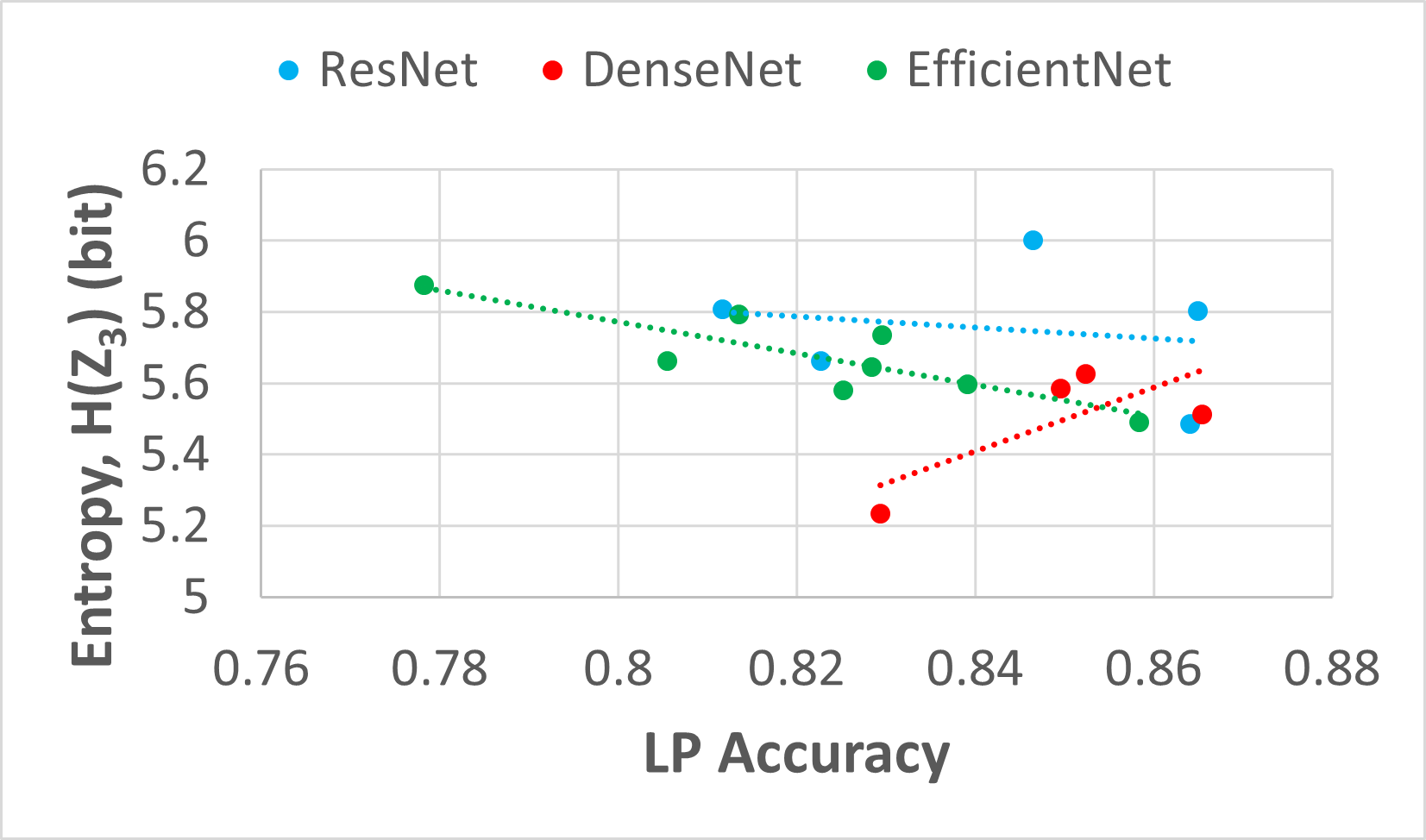}
\caption{Entropy vs linear probe on CIFAR-10, for networks pre-trained on ImageNet.}
\label{fig:sl-entropy-cifar10}
\end{minipage}
\hspace{0.5cm}
\begin{minipage}[b]{0.475\linewidth}
\centering
\includegraphics[width=\textwidth]{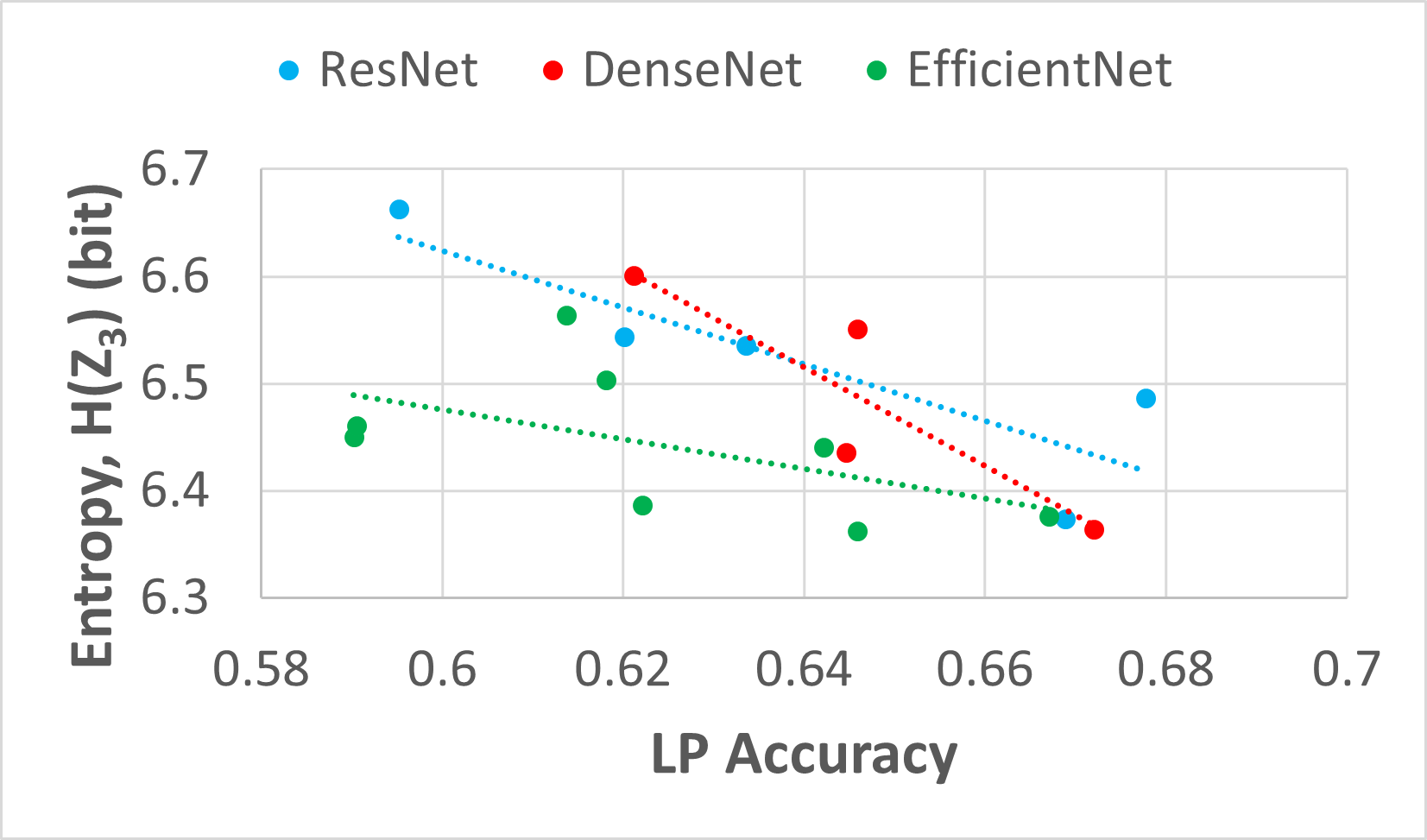}
\caption{Entropy vs linear probe on CIFAR-100, for networks pre-trained on ImageNet.}
\label{fig:sl-entropy-cifar100}
\end{minipage}
\end{figure*}

It is not currently clear what causes these outliers to arise --- whether they are outliers in the raw representational space embeddings $Z_{512}$ as well as in the UMAP-reduced $Z_3$ and if so why these samples deviate from the others.
One possible explanation could be that these images have unusual properties which the network has not yet internalized and hence behave like out-of-domain (OOD) data.
In any case, it appears that for SimCLR and MoCo-v2, the entropy remains low and unstable during the early stages of training, before stabilizing at a higher entropy and then following a linear downward trend.
This may also be the case for SimSiam, but from a high initial entropy to a lower entropy prior to an increasing trend.
Although it is not as visible for CIFAR-10 trials, such behaviour may still exist in the training prior to our first examined milestone.
Lastly, also of note are the apparent drops in entropy around every 100 epochs seen in \autoref{fig:simsiam-entropy-cifar100}. This periodicity appears to be inconsistent and we currently do not believe it to be significant. This behaviour may be caused stochastic elements introduced by the dimensionality reduction process.

Due to the limited number of model examples, all but the strongest correlating metrics did not produce significant results for our pre-trained models. 
Here, entropy is the only metric which produces a statistically significant result for the overall case on pre-trained torchvision models, showing a largely negative correlation with model quality across architectures.
Our results for entropy are presented in \autoref{fig:sl-entropy-cifar10} and \autoref{fig:sl-entropy-cifar100}.
In these figures, we observe that there appears to be a degree of overlapping points between architectures along the same general negative trend, which is a promising sign for a capable cross-architecture comparison metric.
We also note that the weak correlation result for CIFAR-10 appears to be caused by a DenseNet point and a ResNet point.

\section{Conclusion}
The label-free metrics we propose in this paper are generally indicative of the quality of the embedding space as measured with downstream classification when training the network with SimCLR or MoCo-v2.
However, none of our metrics were able to consistently measure the utility of the embedding space learned with SimSiam.
If the cause for why entropy reverses direction depending on methodology can be identified, entropy may be a viable means of label-free learning monitoring or potentially a means to compare different models.
Further work is needed to resolve why the metrics work relatively well only with some SSL methodologies and to test them on additional methods.

As cross-architecture measures of model quality, clustering-based metrics appear to be insufficient as even with ground truth, we were unable to obtain strong results in an overall scenario encompassing ResNet, DenseNet, and EfficientNet architectures. However, there is some evidence that entropy may be architecture-independent given the significant result for CIFAR-100 in this overall case and by visual examination of entropy behaviour compared to LP accuracy. Additional architecture examples would help establish greater confidence in our results.

\FloatBarrier

\section*{Acknowledgment}

This work was supported by the Ocean Frontier Institute as part of the Benthic Ecosystem Mapping and Engagement project.

\bibliographystyle{IEEEtran}
\bibliography{references}

\end{document}